%% file: main.tex
\documentclass{article}
\usepackage{ijcai22}

\usepackage{times}
\usepackage{soul}
\usepackage{url}
\usepackage[hidelinks]{hyperref}
\usepackage[utf8]{inputenc}
\usepackage[small]{caption}
\usepackage{graphicx}
\usepackage{amsmath}
\usepackage{amsthm}
\usepackage{booktabs}
\usepackage{algorithm}
\usepackage{algorithmic}
\usepackage{color}
\usepackage{newfloat}
\usepackage{listings}
\floatstyle{ruled}
\newfloat{listing}{tb}{lst}{}
\floatname{listing}{Listing}
\title{Understanding Narratives through Dimensions of Analogy}

\author{
Thiloshon Nagarajah
\and
Filip Ilievski\And
Jay Pujara
\affiliations
Information Sciences Institute, University of Southern California\\
\emails
\{tnagaraj, ilievski, jpujara\}@isi.edu
}

\begin{document}

\maketitle

\begin{abstract}
Analogical reasoning is a powerful reasoning tool that enables humans to connect two situations, and to generalize their knowledge from familiar to novel situations. Cognitive Science research provides valuable insights into the richness and complexity of analogical reasoning, together with implementations of expressive analogical reasoners with limited scalability. Modern scalable AI techniques with the potential to reason by analogy have focused on the special case of proportional analogy, and have seldom been applied to understand higher-order analogies. In this paper, we aim to bridge the gap on the task of narrative understanding, by: 1) formalizing six dimensions of analogy based on mature insights from Cognitive Science research, 2) annotating a corpus of fables with each of these dimensions, and 3) defining four tasks with increasing complexity that enable scalable evaluation of AI techniques. Experiments with language models and neuro-symbolic AI reasoners on these tasks reveal that state-of-the-art methods can be applied to reason by analogy with a limited success, motivating the need for further research towards comprehensive and scalable analogical reasoning by AI. We make all our code and data available.
\end{abstract}

\input{section/intro}
\input{section/analogies}
\input{section/data}

\input{section/method}
% \input{section/results}
\input{section/relatedwork}

\input{section/conclusion}

\bibliography{paper}
\bibliographystyle{named}

% \section{Acknowledgments}
% AAAI is especially grateful to Peter Patel Schneider for his work in implementing the original aaai.sty file, liberally using the ideas of other style hackers, including Barbara Beeton. We also acknowledge with thanks the work of George Ferguson for his guide to using the style and BibTeX files --- which has been incorporated into this document --- and Hans Guesgen, who provided several timely modifications, as well as the many others who have, from time to time, sent in suggestions on improvements to the AAAI style. We are especially grateful to Francisco Cruz, Marc Pujol-Gonzalez, and Mico Loretan for the improvements to the Bib\TeX{} and \LaTeX{} files made in 2020.

% The preparation of the \LaTeX{} and Bib\TeX{} files that implement these instructions was supported by Schlumberger Palo Alto Research, AT\&T Bell Laboratories, Morgan Kaufmann Publishers, The Live Oak Press, LLC, and AAAI Press. Bibliography style changes were added by Sunil Issar. \verb+\+pubnote was added by J. Scott Penberthy. George Ferguson added support for printing the AAAI copyright slug. Additional changes to aaai22.sty and aaai22.bst have been made by Francisco Cruz, Marc Pujol-Gonzalez, and Mico Loretan.

% \bigskip
% \noindent Thank you for reading these instructions carefully. We look forward to receiving your electronic files!

\end{document}

%% file: section/intro.tex
\section{Introduction}

Understanding narratives requires command of qualitative causal relationships between events, entities, and quantities. Qualitative Reasoning (QR) models have been developed to formalize such relationships in accordance with human mental models~\cite{forbus2011qualitative}, and recent work has demonstrated their ability to model social reasoning~\cite{tomai2007plenty,dehghani2008integrated}. Yet, building scalable cognitive AI architectures requires qualitative representations to be combined with mechanisms that enable efficient and effective generalization and reuse of knowledge, through analogy.

% on core qualitative aspects of common sense, such as time, space, and quantities. Narratives are full with qualitative causal relationships between events, entities, and quantities.

% , and describe how one event or quantity might be impacted by another

% Qualitative reasoning models have long focused on modeling the 

% Qualitative modeling provides techniques for 

% The project will gain understanding of how to develop technology that can connect two situations based on high-level analogical similarities. 
Cognitive Science research dictates that humans develop the ability for analogical reasoning in the first several years of their lives, and rely on this ability to understand, explain, or imagine situations across domains~\cite{holyoak1995mental}. In law, practitioners rely on precedents to make decisions in novel complicated cases. In education, teachers seek familiar situations to capture the student intuition about novel concepts and processes. In politics, leaders craft their arguments based on the lessons learned from impactful events that are commonly known amongst their people. In business, innovators get inspired from the design of processes and objects that have similar function or purpose, being it found in nature or man-made. 

Analogical reasoning can manifest in many ways. Some may serve to help describe an object, for example a fire engine may be red like a tomato. Others may be figurative, comparing a lover to a summer's day. Analogies may draw on relationships between entities, such as comparing the Sun and the Earth to the Earth and the Moon, based on a \textit{revolves} relation. Analogies may also describe events such as comparing a natural disaster to a war, or convey causal information, such as the contributors to the Great Depression and the financial crisis. Given the diversity of these analogies, building AI agents to understand and generate analogies must possess a variety of skills for relational awareness.

% % what this means for AI
% AI agents that support human decision making, creative thinking, and argumentation must possess such analogical reasoning skills. Imagine a teacher that is unsure how to explain the internal processes inside an atom.  An AI agent could then step in and suggest possible analogies that are tailored to the education level of the students. Assuming that the students already understand the workings of the solar system, the AI could suggest the analogy of the solar system, first introduced by Kepler~\cite{}. In this popular analogy, the atom is like the solar system because it is composed of a nucleus which attracts electrons; just like the solar system has the sun which attracts planets. If this analogy is unfamiliar to the class, the AI model could suggest a weaker, more familiar, analogy: the particles inside an atom behave similar to people that play a game of pulling a rope. The same AI model could be used to construct a political argument, by drawing an analogy between Donald Trump's accusation of China being a creator of COVID-19, and a prior statement of Emmanuel Macron in which he accuses Iran of developing nuclear weapons. Both of these situations are accusative statements whose veracity is unclear, made by heads of two Western states, about life-threatening actions of Asian states which are not considered Western allies. 
% \filip{too many tangents, make it more focused - use it to explain desiderata}

% what happens now 
While a rich body of Cognitive Science work on analogical reasoning exists~\cite{gentner1983structure,holyoak1995mental,tversky1977features}, modern AI techniques have not been adapted to this challenge at scale.
As machine learning is the dominant AI paradigm in the past decade, AI models largely base their predictions on prior experiences. However, they generally lack analogical inference mechanisms, i.e., mechanisms to distill high-level patterns, such as entire situations or events, from exemplar data. This prevents neural methods from learning efficiently with little data and understanding, imagining, and explaining novel situations~\cite{chen2019human}. For example, large language models like BERT~\cite{devlin2018bert} and GPT-3~\cite{brown2020language} struggle with understanding analogies between proportions~\cite{ushio-etal-2021-bert}, and they are likely to perform poorly on more complex analogies between two situations or domains. Qualitative representations~\cite{forbus2011qualitative} facilitate the continuous modeling of situations, which is essential for narrative understanding, but they require mechanisms to generalize to a wider set of examples. 
Seminal cognitively inspired architectures~\cite{forbus1995mac,aamodt1994case,Forbus2017AnalogyAQ} are able to draw analogies between situations, yet, adapting them to understand narratives at scale is an open challenge.

% but struggle to grasp the implications of variations in data, and their generalization beyond small hand-crafted datasets has proven difficult.

To bridge this gap between cognitive insights and AI algorithms, we develop a novel taxonomy of six dimensions of analogy. We apply this taxonomy to annotate analogies in pairs of fables, and we perform extensive evaluation and analysis of state-of-the-art AI methods across the different analogical categories. Concretely, our contributions are: % as follows:
\begin{enumerate}
    \item We motivate and formally define six dimensions of analogy that are essential for AI models to understand and reason qualitatively with in situations such as stories. 
    \item We annotate 44 story pairs with each of the six analogical dimensions. We also correlate these annotations with literal similarity and emotional arc similarity.
    \item We define four analogical tasks on this dataset and evaluate several representative state-of-the-art methods based on language models and frame sequences, noting that these are insufficient to capture variants of analogical similarity. %We anticipate that explicit modeling of continuous qualitative aspects of narratives such as time, space, and quantity, holds the promise to improve the narrative understanding by AI through analogy in the future.
\end{enumerate}

% While qualitative modeling of time, space, and quantity are not explicitly represented in this work, we speculate that including such continuous aspects of situations in the future will improve the  see our analogical dimensions

%% file: section/analogies.tex
\section{Dimensions of Analogical Reasoning}

\subsection{Insights from Cognitive Science}

People are born without an analogical reasoning skill, yet, this skill has been shown to develop over time with the expansion of knowledge~\cite{gentner1988metaphor}. 
\citeauthor{holyoak1995mental} specify three levels of analogy, each tied approximately with human age as a proxy for this knowledge: 1) Attributive - A is like B because they share an attribute value (e.g., the ocean is like this car as both as blue). This analogical skill is formed in humans between age 1 and 2~\cite{gentner1986systematicity}. 2) Proportional - A is like B because they have a similar role in their pairwise proportions. For example, car is like sofa because (car-larger than-motorcycle) and (sofa-larger than-chair). This is formed at human age of 3 to 4. 3) Structural - A is like B because their structures align, typically due to causality. Humans acquire structural reasoning at 5 - 6 years of age.

According to~\cite{gentner1997structure}, analogy is drawn between two domains or situations. A successful analogy requires three criteria to be fulfilled~\cite{holyoak1995mental}. 1) Systematicity - the two analogical structures should be aligned - if they are not, comparison is meaningless. 2) Relational similarity - the two structures should have relational similarity, i.e., their fillers should play a similar role in the overall structure of the domain or the situation. 3) Purpose - as the two structures are not overall very similar, drawing an analogy is a creative process that is driven by a certain goal.

Analogy focuses on relational similarity, and as such it differs from other notions of similarity, such as literal similarity, mere appearance, and anomaly~\cite{tversky1977features}. \textit{Literal similarity} between two objects A and B is proportional to the intersection of their features and inversely proportional to the features that differ ($A-B$ and $B-A$), where features include both attributes and relational predicates. \textit{Mere appearance} focuses on similarity between attributes and not the relationships (e.g., moon - coin), while anomaly refers to a comparison with no attribute nor relational overlap~\cite{gentner1983structure}. 

\subsection{Our Taxonomy}

Guided by the rich body of theoretical work, we devise six dimensions of analogy that we expect to be valid for narrative structures. Namely:

\paragraph{1. Shallow Attribute Analogy (SAA)} - Characters in the two situations (here, stories) have explicitly stated \textit{physical} attributes that are similar. For example, given story A where one of the protagonists is a brown fox and story B that features a brown deer, one can draw an analogy between the two stories based on shallow attribute correspondence: brown(fox) $\leftrightarrow$ brown(deer)).
 
\paragraph{2. Deep Attribute Analogy (DAA)} - Characters in the two stories have \textit{abstract} attributes that are similar. For example, given story A where an ass is depicted as naive and story B that features a  naive deer, one can draw an analogy between the two stories based on deep attribute correspondence: naive(ass) $\leftrightarrow$ naive(deer).

\paragraph{3. Relational Analogy (RA)} - Character pairs in the two stories have a similar first-order relationship to each other. For instance, given a story A where a fox and a deer are friends, and story B where a conman is a friend with a countryman, we draw a relational analogy: friendswith(fox,deer) $\leftrightarrow$ friendswith(conman, countryman).
\paragraph{4. Event Analogy (EA)} - Two stories involve the same frame or similar event, e.g., both story A and story B involve a dangerous event.
\paragraph{5. Structural Analogy (SA)} - At least two events, connected with causal links, can be matched across two stories. For instance, the merchant chases an ass and the ass runs as a result of it in story A; a tiger chases a rabbit and the rabbit runs away as a result in story B. Formally: chase(merchant,ass) \& run(ass) \& cause(chase,run) $\leftrightarrow$ chase(tiger,rabbit) \& run(rabbit) \& cause(chase,run).
\paragraph{6. Moral/Purpose (MP)} - Two stories have morals that are compatible, and can be aligned with each other. For instance, the moral of story A ``Know thyself'' corresponds to the moral of story B: ``Know your worth''.

\paragraph{}Suspecting that, unlike analogy, surface similarity is easily captured by state-of-the-art models, we also consider \textbf{Literal Similarity (LS)}, which indicates whether two stories use similar words such as the same character or the same location. For instance, lion in combat with a bull in the jungle $\leftrightarrow$ lion walking with a fox in the jungle.

%% file: section/data.tex
\section{Analogical Reasoning Benchmark}

We collected the primary dataset by scraping Aesop Fables from Litscape.com.\footnote{\url{https://www.litscape.com/indexes/Aesop/Morals.html}} The dataset had 237 fables and corresponding single-sentence moral except for 36 fables without record of morals. Each fable was short with most stories having 3 - 8 sentences and 50 - 150 words. We performed two iterations of annotation to classify single stories and annotate analogical categories for pairs of stories.

\subsection{Single Story Annotation}

\begin{figure}[htp]
    \centering
    \includegraphics[width=8cm]{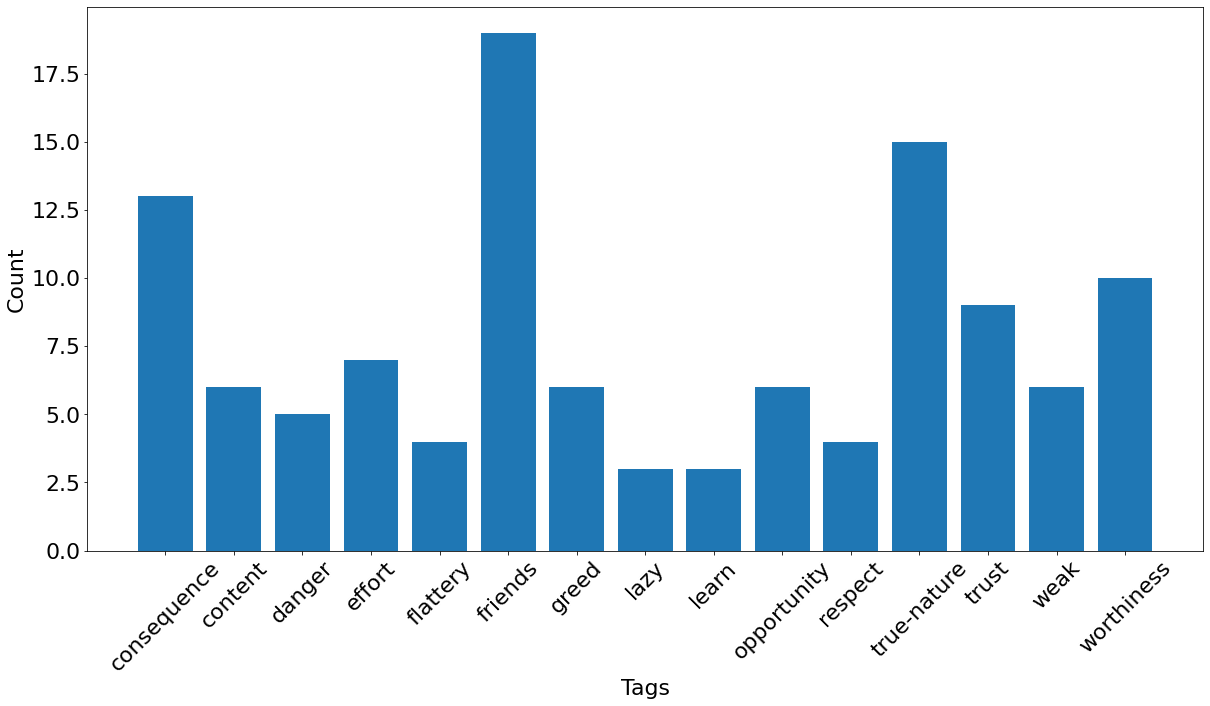}
    \caption{Moral distribution of stories.}
    \label{fig:moral}
\end{figure}

In the first iteration, one annotator manually annotated each fable based on its moral with one of 15 classes. 8 of the classes were taken from the tags defined in the source website, yet most of the fables were not accurately tagged. The annotator introduced 7 more classes such that it best explains the moral. We annotated the moral rather than the story to be able to learn from the moral and generalize to the story. Each story can be annotated with multiple moral classes if they apply. This process resulted in 116 fables with 15 classes: CONSEQUENCE, CONTENT, DANGER, EFFORT, FLATTERY, FRIENDS, GREED, LAZY, LEARN, OPPORTUNITY, RESPECT, TRUE-NATURE, TRUST, WEAK and WORTHINESS. The class distribution of the fables is given in Figure \ref{fig:moral}.
 
\subsection{Story Pairs Annotation}

In the second iteration, we introduced a framework of annotation under six analogy types, described in the previous section. Each fable could have more than one analogy type and each analogy annotation is binary. Besides annotating the binary value (yes/no) for each analogical type, we also included graph-like triples that justify the positive judgment for each analogical type, like \textit{lion - friends with - ass}. This supporting evidence has to be formed as positive triples, negated triples are not accepted. Furthermore, our guidelines dictate that event analogy is a prerequisite for structural analogy since the structural analogy is comprised of at least two analogical events connected with a causal link. When two fables have opposite morals, we consider MP analogy as false. An example of this situation is \textit{1: a merchant gets greedy and kills goose, loosing his wealth}, wherein \textit{2: an ass was content with his food and saved himself from danger}. Both deal with greed, but the first shows the consequence of greed while the second shows a consequence of not being greedy.
The full guidelines can be found online.\footnote{\url{https://docs.google.com/document/d/1OcFD8uSB1HzhYRlR0ubb06wda0daUOBaI7W1axAXryY/edit?usp=sharing}}

To quantify the quality of annotation we computed Inter Annotator Agreement (IAA). We sampled three pairs of stories and three annotators were asked to annotate using the six analogy types. We observed high agreements for SAA and LS, while DAA and SA gave lower agreements. RA, EA and MP were mostly close to random guessing. The primary reason for low agreement scores was that often analogical annotation used implicit attributes or events not included in the short fables, while only explicit annotation would yield very few analogical matches.
This signals a need for further discussion and specification of the annotation guidelines and training. The Cohen Kappa IAA score between annotators 1, 2, and 3 is shown in Table~\ref{tab:cohen}.

\begin{table}
\centering
\caption{Cohen Kappa IAA Scores.}
\label{tab:cohen}
\scriptsize
\begin{tabular}{||p{1cm} | p{0.5cm} p{0.5cm} p{0.5cm} p{0.5cm} p{0.5cm} p{0.5cm} p{0.5cm}||} 
 \hline
 Annotator  & SAA & DAA & RA & EA & SA & MP & LS \\ 
 \hline\hline
 
  1 VS 2 & 1.00 & -0.5 & 0.39 & 1.00 & 0.00 & 1.00 & 0.39 \\  
 \hline
  1 VS 3 & 1.00 & 1.00 & 0.00 & 0.00 & 0.00 & 0.00 & 0.39 \\ 
 \hline
  2 VS 3 & 1.00 & -0.5 & 0.00 & 0.00 & -0.5 & 0.00 & 1.00 \\
 \hline \hline
\end{tabular}
\end{table}

%% file: section/method.tex
\section{Methods}

We identified four main tasks to learn or evaluate analogies with the annotations. Each task is built from the results and understanding from the previous tasks and makes use of various characteristics of the annotation process. These tasks range from exploratory data analysis to using transfer learning to assess generality, which provide a robust pipeline for future research and serve as the initial benchmark. In task 1, we model the high level semantic themes of the fables by using their morals, and learn to cluster these thematic analogies to their tags. In task 2,  we use different techniques using lexical and semantic structures of fables to generate candidate analogical fable pairs and evaluate the efficacy of these generative methods. Next, in task 3, we evaluate different learning approaches to identify the specific analogy types within pairs. Finally, in task 4, we evaluate the generality of our methods by trying to transfer the learned analogical relationship from a different dataset from a potentially unrelated domain. We present the task details, data, models, and evaluation in detail in the following subsections.

\subsection{Task 1: Moral Clustering}

\textbf{Task description:} At the most abstract level, analogical reasoning can be approximated by grouping similar items together. We approach this setting by clustering fables to their respective high-level morals. If two stories are about a fox being greedy and a merchant being greedy, first step of deriving this analogy is to identify that both stories belong to the moral tag GREED. Since morals in the dataset are one sentence summary of the fables, we experimented with methods to cluster the morals to one of 15 classes. To build these methods we used different characteristics of the story-moral pairs as below:

\textbf{Language model baseline:} Recent breakthroughs in NLP use contextual-language models to produce embeddings that represent documents. Word level language models have shown the ability to identify analogies like \textit{man:woman::king:queen}, but work on sentence and document level modeling has mostly focused on NLI or paraphrase detection. 
We finetuned a RoBERTa \cite{DBLP:journals/corr/abs-1907-11692} model on the morals and their tags, and retrieved [CLS] hidden layer of the morals as a sentence representation. This yielded poor performance as the number of classes were too large relative to the size of the dataset and some classes had very few training instances. As an alternative, we experimented with one-vs-one and one-vs-all classifiers. Embedded morals were passed to a 3-layer neural network with softmax output using a cross entropy loss function. We observed a low evaluation accuracy, suggesting such models are too complex for our limited data. Finally we built 15 one-vs-all logistic regression classifiers for each tag which performed better than neural network. After 100 iterations of under sampled one-vs-all classifiers resulted in accuracies ranging from 59\% to 64\% but a low F1 score of 0.3. Manual inspection of the variety of the moral tags suggest these tags did not adequately represent fables, leading to a different methodology for subsequent tasks.

\textbf{Frame baseline:}
 We observed that the stories under certain tags have similar common frames or frame sequences. For example, fables under FRIENDSHIP have \textit{Travel-Collaboration} and fables under TRUST have \textit{Manipulation} as common frames.  We also noticed that stories tend to follow certain similar sequence of actions which could be captured with a sequence model. But to generalize the types of these actions, we choose to use frames parsed from FrameNet \cite{BakerFillmoreLowe:98} using the open-sesame frame parser \cite{https://doi.org/10.48550/arxiv.1706.09528}. To answer the question whether the sequence of frames or even existence of frames could differentiate the type of the fable, we built different methods to classify morals. The simplest notion was to consider the number of occurrence of a frame and bi-grams of frames as numeric feature and learn with a neural network model. Then we considered sequence of frames embedded with RoBERTa as input for a Bi-LSTM model \cite{10.1162/neco.1997.9.8.1735}. We considered 376 frames found in all the stories as well as only the most common 15 frames per tag as features. 

 \textbf{Results:}
 The neural network classifier to predict one of 15 tags from frame count performed poorly. When all 376 frames were considered as network features, accuracy of 0.10 was attained. Using only the most common 15 frames increased accuracy slightly to 0.12. With under-sampled data, an one-vs-all classifier performed better than random guessing though F1 score was still low (0.35). Accuracies of the selected one-vs-all classifiers are given in Table \ref{tab:onevsall}. We leave it to subsequent work to analyze to what extent are the obtained results influenced by noise in the frame parser, noise in the annotations, or the usefulness of the frame representations.

\begin{table}
\begin{center}
\caption{Accuracies of one-vs-all neural classifiers.}
\label{tab:onevsall}
\begin{tabular}{||c c||} 
 \hline
 Tag & Accuracy \\ 
 \hline\hline
 Consequence & 0.64 \\ 
 \hline
 Content & 0.64 \\
 \hline
 Danger & 0.62 \\
 \hline
 Effort & 0.51 \\
 \hline
 Flattery & 0.59 \\
 \hline
 Friends & 0.54 \\ 
 \hline
 Greed & 0.64 \\ 
 \hline
 Opportunity & 0.59 \\ 
 \hline
\end{tabular}
\end{center}
\end{table}

\begin{table}[!t]
\centering
\caption{Analogies of story pairs.}
\label{tab:correlation}
\scriptsize
\begin{tabular}{||p{1.8cm} | p{1cm} | p{0.7cm} p{0.7cm} p{0.7cm} p{0.7cm}||} 
 \hline
   & maximum & Semantic & Frames & Lexical & Random \\ 
 \hline\hline
 Story count & 44 & 9 & 10 & 12 & 13 \\ 
 \hline
 Method average & 6 & 2.22 & 1.90 & \textbf{3.00} & 1.54 \\ 
 \hline \hline
 SAA & 1 & 0.00 & 0.00 & 0.08 & 0.08 \\
 \hline
 DAA & 1 & \textbf{0.44} & \textbf{0.60} & \textbf{0.58} & 0.31 \\
 \hline
 RA & 1 & \textbf{0.56} & \textbf{0.40} & \textbf{0.58} & \textbf{0.54}  \\
 \hline
 SA & 1 & 0.22 & 0.10 & 0.25 & 0.00  \\
 \hline
 MP & 1 & 0.11 & 0.10 & \textbf{0.42} & 0.23  \\
 \hline
 LS & 1 & 0.33 & 0.00 & \textbf{0.50} & 0.15  \\
 \hline \hline
 SSS & 1 & \textbf{0.56} & \textbf{0.70} & \textbf{0.58} & 0.23  \\

 \hline
\end{tabular}
\end{table}

\subsection{Task 2: Analogical Pair Generation}

\textbf{Task description:}
Our analogy framework was built to compare analogies in pairs of stories. But selection of stories to be paired cannot be done arbitrarily. The rationale behind choosing a pair of story has to be evaluated to be able to understand the implications of the method. Different sampling methods could affect the ability to learn analogy from fables. For instance, lexical methods could find the fables that share similar words, yet might not be a great method to find analogies. A semantic or structure bases method could do well with identifying the similar buildup of stories which does not necessarily need to have similar lexical features. In this task, we define the different ways pairs could be created and methods to evaluate their efficacy. 

\textbf{Language model baselines:} To build story pairs, we experimented with different sampling techniques under lexical, semantic, or structural methods. We noticed that certain stories have same characters and same interactions between characters, which could be captured well with lexical methods. But since we are more interested in the implicit analogies in the story rather than the explicitly stated similar words, we also considered semantic and frame similarities. For each story, we picked the most similar story using these methods and generated 348 story pairs. We added another 348 pairs of stories that were randomly selected to a final total of 696 story pairs. Implemented sampling techniques were as follows: 

\begin{enumerate}
  \item Lexical - We converted each word in the fable to Word2Vec \cite{https://doi.org/10.48550/arxiv.1301.3781} embedding and calculated the weighted average of the story with TF-IDF. We used cosine similarity between these story-level embedding to find the most matching fable pairs. 
  \item Semantic - We converted each fable to a RoBERTa \cite{DBLP:journals/corr/abs-1907-11692} embedding and retrieved the embedding value for the [CLS] token. We used cosine similarity on top of this embedding.  
  \item Frame - We parsed frames in each fable with open-sesame and calculated sentence edit distance between the frames to find the most similar frame sequence. Sentence edit distance is adopted from word edit distance algorithm by considering each word as a character in the original algorithm.  Frame edit distance was scaled to fit different frame lengths. To have high variance, we kept only frames found in story 1 as features for edit distance.
 \end{enumerate}

\begin{figure}[!t]
    \centering
    \includegraphics[width=8cm]{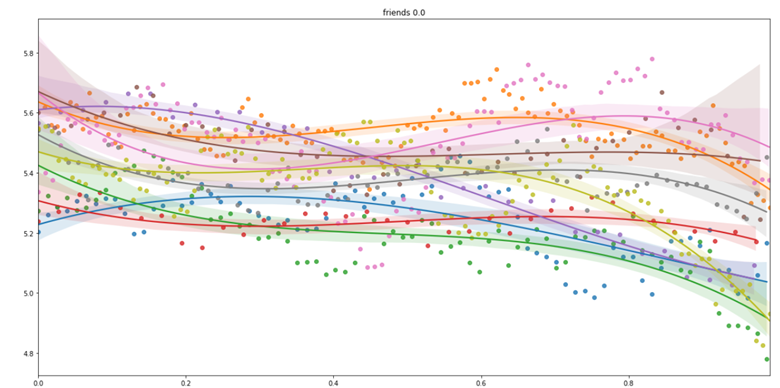}
    \caption{Fables with a story shape High - High - Low.}
    \label{fig:shapes}
\end{figure}
 
 \textbf{Story shape baseline:}
 In addition to language models, we considered story shapes \cite{DBLP:journals/corr/ReaganMKDD16} to generate story pairs. Story shapes are arcs generated from sentiment scores of the fable as the story progresses. Though these arcs do not explicitly provide the information about the plot of stories, they give an overall characteristic curve of the emotions in the story. Table~\ref{tab:shapes} lists the prototypical story shapes identified by \citeauthor{DBLP:journals/corr/ReaganMKDD16}. Using the definition of these curves, we can tag the stories which have similar arcs as similar stories. We analysed how emotions in the story change over the duration of the story to classify the story into one of five common story shapes, using the arc types in prior work. To calculate emotional changes, we used the MTLabs \cite{Dodds_2011} vocabulary to assign happiness scores (hedonometric score) for each word in the story. A sliding window of 30 words was used to calculate the average of the happiness within the window and averages throughout the story was recorded. The HIGH or LOW points are calculated as average of the beginning (first 30\% of the story), middle (middle 40\%) and end (last 30\%) of the story. The averages of the 30-word window typically lie between a hedonometric score of 5 to 7 \cite{10.1371/journal.pone.0026752} and for fables the average of all the stories was 5.4. Based on the assumption 5.4 denotes a neutral tone, HIGH was taken as an average of above 5.4+0.2 and LOW as an average less than 5.4-0.2.

 \textbf{Results:} We evaluated the strength of the 3 language model based sampling methods by assigning 0 or 1 for analogies and summing up over the methods. The scores are presented in table \ref{tab:correlation}. The Word2Vec-based lexical method was found to have the highest analogy score overall. Frames based method gives high analogical similarity for DAA. The MP similarity, which is the most important analogy for this research was high with lexical sampling. Even the highest analogical scores were between 0.4 and 0.6, demonstrating that the most similar stories under different sampling methods do not necessarily share comparable analogies. It is also worth noting that RA and DAA was comparatively high even for the random sampling method, showing how any fable could be justified to have an DAA or RA analogy mainly due the nature of the fable stories. This cautions of a possible high annotator bias that needs to be addressed with strict and elaborate annotation guidelines.

\begin{figure}[!t]
    \centering
    \includegraphics[width=8cm]{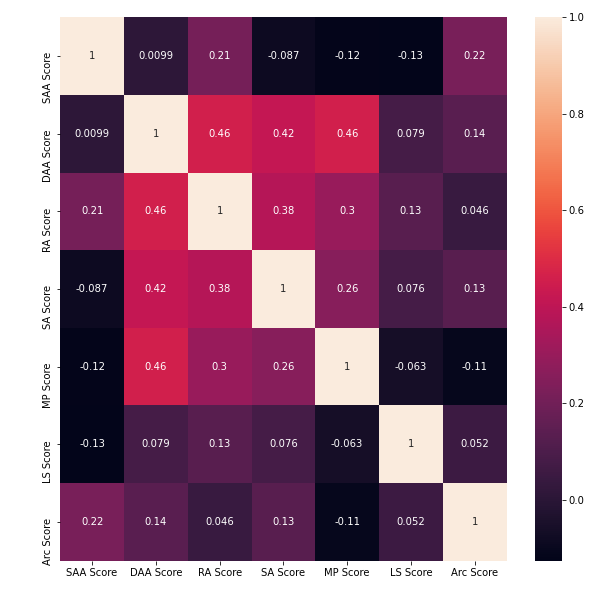}
    \caption{Correlation of analogies}
    \label{fig:correlation}
\end{figure}

We further analyzed how the analogy types are correlated with each other and presented in Figure \ref{fig:correlation}. We observed that MP is mostly correlated with DAA and that DAA was also highly correlated with SA and RA. We note that the correlation scores were only as high as 0.46 and most of the correlations were under 0.2, showing how the analogies are not as correlated as one would expect to be.

\begin{table}
\caption{Exemplar story shapes.}
\label{tab:shapes}
\begin{tabular}{||c c||} 
 \hline
 Fable Arc & Arc type \\ 
 \hline\hline
 High - High - Low (Figure \ref{fig:shapes}) & Tragedy \\ 
 \hline
 Low - Middle - High & Rags to Riches \\
 \hline
 High - Low - High & Cinderella \\
 \hline
 Low - Middle - High & Rags to Riches \\
 \hline
 Low - Middle - Low & Oedipus \\
 \hline
\end{tabular}
\end{table}

 The classifier to learn from story shapes yielded poor accuracy showing story arcs alone cannot be used to identify similar stories.
 
\subsection{Task 3: Analogy Type Prediction}

\textbf{Task description:} Building up from Task 2 of generating similar story pairs, the next natural step is to be able to predict analogy types. We could learn from the pairs of stories generated by different sampling methods to predict whether an analogical dimension holds for a story pair or not.

\textbf{Baseline:} We tuned a cross-encoder DistilRoBERTa-base \cite{Sanh2019DistilBERTAD} model to identify analogies. We formulated the problem as binary classification and built 7 classifiers: one for each of the six analogy types and one for literal similarity. 

\textbf{Results:} Table \ref{tab:analogypred} shows the obtained results together with the ration of the positive class for each analogical class. We also report F1-scores to account for the imbalance of certain classes. The cross encoder obtained high F1 performance for SAA, DAA, and RA. Yet, it performed only slightly better than random guessing for other analogy types. For analogy types with a balanced class distribution, the classifier was able to result in high F1 score, and analogy types with skewed classes performed poorly. Since this experiment was performed only on 44 annotated pairs, we are optimistic that having a larger dataset may yield better performing models.

\begin{table}
\begin{center}
\caption{Performance of Analogy Type Classifiers.}
\label{tab:analogypred}
\begin{tabular}{||c c c c||} 
 \hline
 Analogy & Positive Class Ratio & Accuracy & F1 \\ 
 \hline\hline
 SAA & 0.05 & 1.0 & 1.0 \\
 \hline
 DAA & 0.42 & 0.77 & 0.67\\ 
 \hline
 RA & 0.51 & 0.77 & 0.77 \\
 \hline
 SA & 0.08 & 0.66 & 0.4\\
 \hline
 MP & 0.22  & 0.77 & 0.43\\
 \hline
 LS & 0.28  & 0.88 & 0.47\\
 \hline
\end{tabular}
\end{center}
\end{table}

Below we provide an example of a story pair that has Deep Attributive, Relational, and Event analogies as identified by our classifier. Both soldier in story 1 and Hen in story 2 are caring, which is a deep attributive analogy. Their care and act of caring for horse or egg is relational and event analogies. These stories does not share a moral purpose as in story 1 the moral would be that \textit{fickle care causes waste}, wherein story 2 it is to \textit{be careful who you care for}. 

\begin{quotation}
\scriptsize
Story 1: A HORSE SOLDIER took great pains with his charger. As long as the war lasted, he looked upon him as his fellow-helper in all emergencies, and fed him carefully with hay and corn. But when the war was over, he only allowed him chaff to eat and made him carry heavy loads of wood, and subjected him to much slavish drudgery and ill-treatment. War, however, being again proclaimed, the Soldier put on his charger its military trappings, and mounted, being clad in his heavy coat of mail. The Horse fell down straightway under the weight, no longer equal to the burden, and said to his master: You must now e'en go to the war on foot, for you have transformed me from a Horse into an Ass.

Story 2: A HEN finding the eggs of a viper, and carefully keeping them warm, nourished them into life. A Swallow, observing what she had done, said: You silly creature! Why have you hatched these vipers, which, when they shall have grown, will surely inflict injury on all of us, beginning with yourself?
\end{quotation}

\subsection{Task 4: Analogical Transfer Learning}

\textbf{Task description:} To evaluate the robustness of our methods, we design a transfer learning experiment, where the objective is to build a baseline that can learn analogies from one dataset and be able to sufficiently transfer that knowledge to a different dataset. 

\textbf{Baselines:} We experimented with the following pipelines:
\begin{enumerate}
    \item Fable2Fable - We used the 116 fables with their tags from our first annotation iteration  to build a corpus of 544 story pairs whose morals belong to the same tag. To created a balanced dataset, we added 544 more story pairs that do not belong to same tag as negative samples. We finetuned a pre-trained cross encoder DistilRoBERTa model to predict whether the story pair belongs to the same tag. We evaluated the model on the 44 story pairs annotated with analogy types by predicting if the pair of story is analogical for each of the seven types.
    \item AILA2Fable - We used the AILA \cite{DBLP:conf/fire/BhattacharyaG0019a} dataset which contains legal cases and their precedent cases to build a positive and negative pairs. We identified 640 pairs of similar cases as positive samples and added another 640 unrelated cases as negative samples. As the legal cases are lengthy, we sampled 500 words from the middle of the document to support the Transformer-based cross encoder. The remaining training and evaluation setup was same as in Fable2Fable.  
\end{enumerate}

\begin{figure}[!t]
    \centering
    \includegraphics[width=7.8cm]{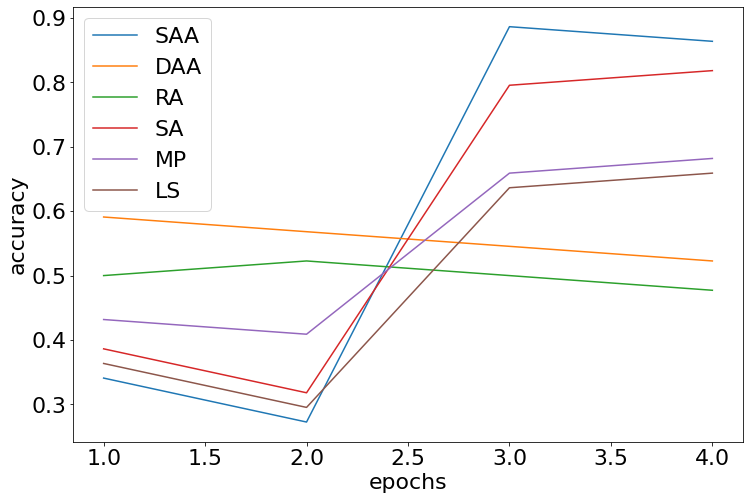}
    \caption{Accuracy of Fable2Fable.}%\filip{replace pipeline 1 with a descriptive name. }}
    \label{fig:accuracy-fable}
\end{figure}

\begin{figure}[!t]
    \centering
    \includegraphics[width=7.8cm]{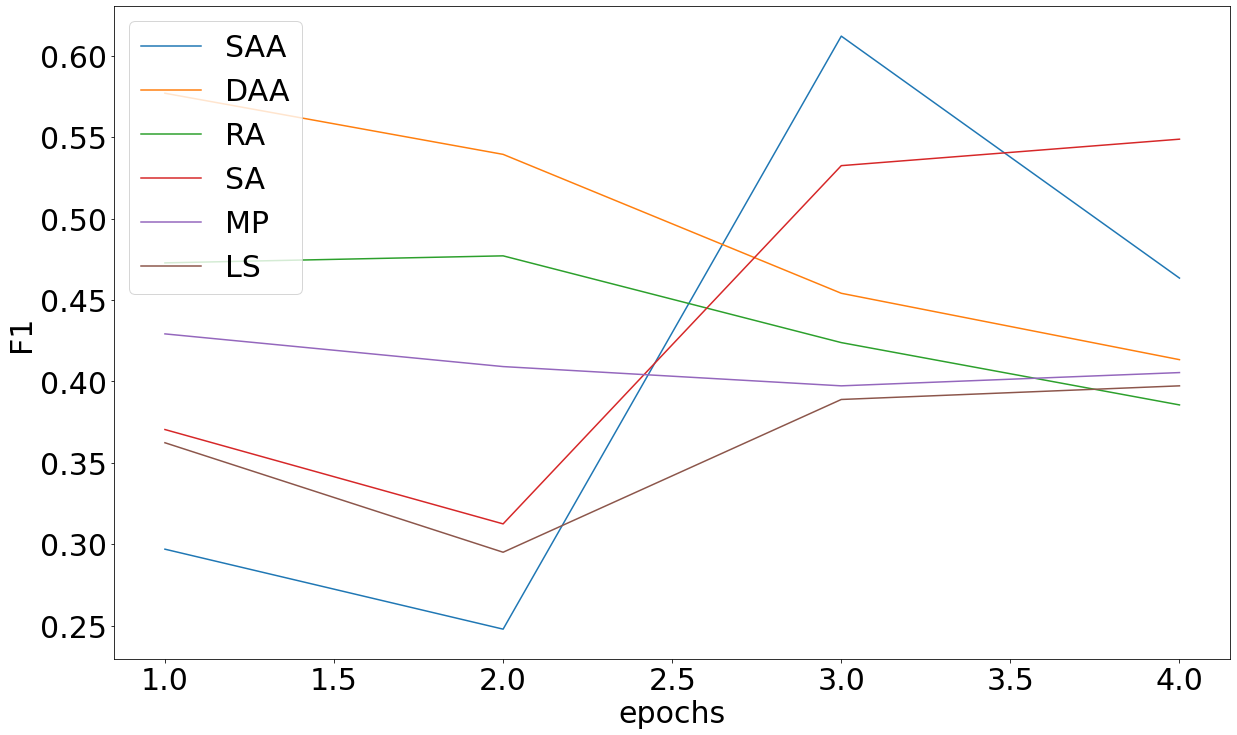}
    \caption{F1 score of Fable2Fable.}%\filip{why different metrics in figure 4 and 5.}}% \filip{replace pipeline 1 with a descriptive name. }}
    \label{fig:f1-fable}
\end{figure}

\textbf{Results:} Fable2Fable was built using DistilRoBERTa \cite{Sanh2019DistilBERTAD} cross encoder and we observed a substantial prediction power in SAA, SA, MP, and LS, with SAA reaching a highest accuracy of 88\%, and SA reaching the highest F1-score of 55\%. Curiously, we observed that the model predictions changed dramatically based on its training over the epochs. The changes in accuracy and F1 scores across the epochs are given in Figures \ref{fig:accuracy-fable} and \ref{fig:f1-fable}, respectively. We note that SAA, SA, MP and LS performed better with more epochs, the performance of predicting DAA and RA analogy was affected negatively. One possible explanation for this different behavior across classes may lie in the class distributions, as DAA and RA are almost perfectly balanced whereas SA, MP and LS have low positive class ratios of 0.1, 0.2 and 0.23, respectively. 

For the AILA2Fable pipeline, DistilRoBERTa was not able to learn sufficiently and we used a pre-trained model that was built for the paraphrasing task (paraphrase-MiniLM-L6-v2). We omit the results from this experiment, as the accuracy and the F1-score per class were close to random guessing. This signals that, while transferring labels for the same domain can be achieved to some extent with pre-trained language models, generalization from an unrelated domain is more difficult and may require a different architecture.

% demonstrating how it is hard to learn from unrelated domain to be applied on fable dataset.

%% file: section/relatedwork.tex
\section{Related Work}

\paragraph{Analogies in cognitive science} Cognitive systems typically consist of multiple phases: retrieval, mapping, abstraction, and representation~\cite{gentner2011computational}. The structural mapping engine~\cite{falkenhainer1989structure,Forbus2017ExtendingST} maps two structures to each other by combining local with structural similarity, ultimately being able to find multiple analogies for a given situation. Similarly, the MAC-FAC  (Many Are Called but Few Are Chosen)~\cite{forbus1995mac} system starts with a cheap retrieval of possible analogies, followed by a a structure mapping step. 
The Companion cognitive architecture~\cite{Forbus2017AnalogyAQ}, which leverages a subset of CYC~\cite{lenat1995cyc}, strives to capture analogical and qualitative reasoning, and has been proven useful for textbook problem solving, moral decision-making, and commonsense reasoning tasks.
Case-based reasoning (CBR) builds novel strategies based on prior experience through four steps: retrieval of similar cases, their reuse into a proposed solution, revision of this solution, and retainment of the updated solution in the collection of prior cases~\cite{aamodt1994case}. Unlike analogical reasoning, CBR is not meant to be used broadly, and it is typically meant to perform a specific task well, thus trading generality for efficiency or other performance measures~\cite{de2005retrieval}. 
Recent work has used case-based reasoning to answer knowledge base queries, by retrieving subgraphs that correspond to structurally similar queries and reasoning over them~\cite{das2022knowledge}. The development of these cognitively inspired architectures is a promising complementary direction. While it is currently unclear how to apply these methods on tasks based on natural language narratives such as the ones in our work, we expect that future research may be able to bridge this gap.

\paragraph{Analogies in AI and NLP}
Several benchmarks for analogical reasoning exist in AI and NLP, e.g., based on psychometric analogy tests~\cite{mikolov2013linguistic,gladkova2016analogy}. These benchmarks focus on capturing the pairwise relations between two concepts (e.g., king - man), sometimes exemplified through the relation between another pair of concepts (e.g., queen - woman). Relational analogies can be learned to some extent from data, as shown
% There have been recent attempts to learn word-level analogies from different data sources. 
in a recent study \cite{DBLP:journals/corr/abs-2105-04949}, where BERT, RoBERTa, and GPT-3 have been evaluated on linguistic and cognitive assessment datasets equivalent to analogy problems on the Scholastic Aptitude Test (SAT). 
% Unlike our attempt to learn analogy from story pairs, this study focuses on single word relationships like word:language to note:music. Though these models learned relational knowledge to certain extend, there was room for several improvements as the result was highly sensitive to model architectures and hyperparameters. 
% As indicated in recent work, 
Structural analogies have been rarely employed in recent AI and NLP work~\cite{lu2019emergence}. \citeauthor{inproceedings}~used frames from FrameNet to build a corpus with paraphrased Aesop fables by using static patterns. While such paraphrased stories can be seen as analogical, we expect that paraphrased story pairs can be easily picked up by latest language models. Efforts that combine analogical reasoning and language modeling exist~\cite{ribeiro2021combining}, yet, it is unclear how to adapt them to the task of understanding narratives. In our study, we focus on six dimensions of analogy, thus targeting high-level structural and relational similarity, and explore how language models and semantic frames can come together to understand narratives.

%% file: section/conclusion.tex
\section{Conclusions}

This paper presented a framework for analogical reasoning by AI techniques. Guided by Cognitive Science theories, we devised six analogical dimensions with increasing complexity: shallow and deep attribute analogy, relational analogy, event analogy, structural analogy, and moral/purpose, and we considered literal similarity as a baseline dimension. We annotated fable narratives with moral tags and the existence of pairwise story analogy, resulting in a small comprehensive evaluation corpus.
We defined four tasks over this data: clustering morals, analogical pair generation, analogical type prediction, and analogical transfer, each revolving around various analogical dimensions.
We evaluated several representative AI techniques based on language models and semantic frames on these four tasks, observing that these techniques are applicable across tasks, yet, their ability to reason by analogy is limited, especially when this requires understanding of complex structures. We make all code and data available to facilitate subsequent work on analogical reasoning in AI: \url{https://github.com/usc-isi-i2/analogical-transfer-learning}.

Based on our results, we anticipate that explicit modeling of continuous qualitative aspects of narratives such as time, space, and quantity, holds the promise to improve the narrative understanding by AI through analogy in the future. Semantic natural language systems that automatically produce qualitative descriptions~\cite{friedman2021extracting} provide an intuitive opening to integrate qualitative representations with analogical reasoning models. Future work should also develop scalable structure-aware methods for analogical reasoning, possibly by building on prior cognitive architectures. Finally, future work should aim to create an evaluation dataset that is orders of magnitude larger, and investigate clever ways to acquire large-scale training data for model development.

%% file: main.bbl
\begin{thebibliography}{}

\bibitem[\protect\citeauthoryear{Aamodt and Plaza}{1994}]{aamodt1994case}
Agnar Aamodt and Enric Plaza.
\newblock Case-based reasoning: Foundational issues, methodological variations,
  and system approaches.
\newblock {\em AI communications}, 7(1):39--59, 1994.

\bibitem[\protect\citeauthoryear{Baker \bgroup \em et al.\egroup
  }{1998}]{BakerFillmoreLowe:98}
Collin~F. Baker, Charles~J. Fillmore, and John~B. Lowe.
\newblock The {B}erkeley {F}rame{N}et project.
\newblock In {\em COLING-ACL {\textquoteright}98: Proceedings of the
  Conference}, pages 86--90, Montreal, Canada, 1998.

\bibitem[\protect\citeauthoryear{Bhattacharya \bgroup \em et al.\egroup
  }{2019}]{DBLP:conf/fire/BhattacharyaG0019a}
Paheli Bhattacharya, Kripabandhu Ghosh, Saptarshi Ghosh, Arindam Pal, Parth
  Mehta, Arnab Bhattacharya, and Prasenjit Majumder.
\newblock Overview of the {FIRE} 2019 {AILA} track: Artificial intelligence for
  legal assistance.
\newblock In Parth Mehta, Paolo Rosso, Prasenjit Majumder, and Mandar Mitra,
  editors, {\em Working Notes of {FIRE} 2019 - Forum for Information Retrieval
  Evaluation, Kolkata, India, December 12-15, 2019}, volume 2517 of {\em {CEUR}
  Workshop Proceedings}, pages 1--12. CEUR-WS.org, 2019.

\bibitem[\protect\citeauthoryear{Brown \bgroup \em et al.\egroup
  }{2020}]{brown2020language}
Tom Brown, Benjamin Mann, Nick Ryder, Melanie Subbiah, Jared~D Kaplan, Prafulla
  Dhariwal, Arvind Neelakantan, Pranav Shyam, Girish Sastry, Amanda Askell,
  et~al.
\newblock Language models are few-shot learners.
\newblock {\em Advances in neural information processing systems},
  33:1877--1901, 2020.

\bibitem[\protect\citeauthoryear{Chen \bgroup \em et al.\egroup
  }{2019}]{chen2019human}
Kezhen Chen, Irina Rabkina, Matthew~D McLure, and Kenneth~D Forbus.
\newblock Human-like sketch object recognition via analogical learning.
\newblock In {\em Proceedings of the AAAI Conference on Artificial
  Intelligence}, volume~33, pages 1336--1343, 2019.

\bibitem[\protect\citeauthoryear{Das \bgroup \em et al.\egroup
  }{2022}]{das2022knowledge}
Rajarshi Das, Ameya Godbole, Ankita Naik, Elliot Tower, Robin Jia, Manzil
  Zaheer, Hannaneh Hajishirzi, and Andrew McCallum.
\newblock Knowledge base question answering by case-based reasoning over
  subgraphs.
\newblock {\em arXiv preprint arXiv:2202.10610}, 2022.

\bibitem[\protect\citeauthoryear{De~Mantaras \bgroup \em et al.\egroup
  }{2005}]{de2005retrieval}
Ramon~Lopez De~Mantaras, David McSherry, Derek Bridge, David Leake, Barry
  Smyth, Susan Craw, Boi Faltings, Mary~Lou Maher, MICHAEL T~COX, Kenneth
  Forbus, et~al.
\newblock Retrieval, reuse, revision and retention in case-based reasoning.
\newblock {\em The Knowledge Engineering Review}, 20(3):215--240, 2005.

\bibitem[\protect\citeauthoryear{Dehghani \bgroup \em et al.\egroup
  }{2008}]{dehghani2008integrated}
Morteza Dehghani, Emmett Tomai, Kenneth~D Forbus, and Matthew Klenk.
\newblock An integrated reasoning approach to moral decision-making.
\newblock In {\em AAAI}, pages 1280--1286, 2008.

\bibitem[\protect\citeauthoryear{Devlin \bgroup \em et al.\egroup
  }{2018}]{devlin2018bert}
Jacob Devlin, Ming-Wei Chang, Kenton Lee, and Kristina Toutanova.
\newblock Bert: Pre-training of deep bidirectional transformers for language
  understanding.
\newblock {\em arXiv preprint arXiv:1810.04805}, 2018.

\bibitem[\protect\citeauthoryear{Dodds \bgroup \em et al.\egroup
  }{2011a}]{Dodds_2011}
Peter~Sheridan Dodds, Kameron~Decker Harris, Isabel~M. Kloumann, Catherine~A.
  Bliss, and Christopher~M. Danforth.
\newblock Temporal patterns of happiness and information in a global social
  network: Hedonometrics and twitter.
\newblock {\em {PLoS} {ONE}}, 6(12):e26752, dec 2011.

\bibitem[\protect\citeauthoryear{Dodds \bgroup \em et al.\egroup
  }{2011b}]{10.1371/journal.pone.0026752}
Peter~Sheridan Dodds, Kameron~Decker Harris, Isabel~M. Kloumann, Catherine~A.
  Bliss, and Christopher~M. Danforth.
\newblock Temporal patterns of happiness and information in a global social
  network: Hedonometrics and twitter.
\newblock {\em PLOS ONE}, 6(12):1--1, 12 2011.

\bibitem[\protect\citeauthoryear{Elson and McKeown}{2009}]{inproceedings}
David Elson and Kathleen McKeown.
\newblock Extending and evaluating a platform for story understanding.
\newblock pages 32--35, 01 2009.

\bibitem[\protect\citeauthoryear{Falkenhainer \bgroup \em et al.\egroup
  }{1989}]{falkenhainer1989structure}
Brian Falkenhainer, Kenneth~D Forbus, and Dedre Gentner.
\newblock The structure-mapping engine: Algorithm and examples.
\newblock {\em Artificial intelligence}, 41(1):1--63, 1989.

\bibitem[\protect\citeauthoryear{Forbus and
  Hinrichs}{2017}]{Forbus2017AnalogyAQ}
Kenneth~D. Forbus and Thomas~R. Hinrichs.
\newblock Analogy and qualitative representations in the companion cognitive
  architecture.
\newblock 2017.

\bibitem[\protect\citeauthoryear{Forbus \bgroup \em et al.\egroup
  }{1995}]{forbus1995mac}
Kenneth~D Forbus, Dedre Gentner, and Keith Law.
\newblock Mac/fac: A model of similarity-based retrieval.
\newblock {\em Cognitive science}, 19(2):141--205, 1995.

\bibitem[\protect\citeauthoryear{Forbus \bgroup \em et al.\egroup
  }{2017}]{Forbus2017ExtendingST}
Kenneth~D. Forbus, Ronald~W. Ferguson, Andrew~M. Lovett, and Dedre Gentner.
\newblock Extending sme to handle large-scale cognitive modeling.
\newblock {\em Cognitive science}, 41 5:1152--1201, 2017.

\bibitem[\protect\citeauthoryear{Forbus}{2011}]{forbus2011qualitative}
Kenneth~D Forbus.
\newblock Qualitative modeling.
\newblock {\em Wiley Interdisciplinary Reviews: Cognitive Science},
  2(4):374--391, 2011.

\bibitem[\protect\citeauthoryear{Friedman \bgroup \em et al.\egroup
  }{2021}]{friedman2021extracting}
Scott~E Friedman, Ian~H Magnusson, and Sonja~M Schmer-Galunder.
\newblock Extracting qualitative causal structure with transformer-based nlp.
\newblock {\em arXiv preprint arXiv:2108.13304}, 2021.

\bibitem[\protect\citeauthoryear{Gentner and
  Forbus}{2011}]{gentner2011computational}
Dedre Gentner and Kenneth~D Forbus.
\newblock Computational models of analogy.
\newblock {\em Wiley interdisciplinary reviews: cognitive science},
  2(3):266--276, 2011.

\bibitem[\protect\citeauthoryear{Gentner and
  Markman}{1997}]{gentner1997structure}
Dedre Gentner and Arthur~B Markman.
\newblock Structure mapping in analogy and similarity.
\newblock {\em American psychologist}, 52(1):45, 1997.

\bibitem[\protect\citeauthoryear{Gentner and
  Toupin}{1986}]{gentner1986systematicity}
Dedre Gentner and Cecile Toupin.
\newblock Systematicity and surface similarity in the development of analogy.
\newblock {\em Cognitive science}, 10(3):277--300, 1986.

\bibitem[\protect\citeauthoryear{Gentner}{1983}]{gentner1983structure}
Dedre Gentner.
\newblock Structure-mapping: A theoretical framework for analogy.
\newblock {\em Cognitive science}, 7(2):155--170, 1983.

\bibitem[\protect\citeauthoryear{Gentner}{1988}]{gentner1988metaphor}
Dedre Gentner.
\newblock Metaphor as structure mapping: The relational shift.
\newblock {\em Child development}, pages 47--59, 1988.

\bibitem[\protect\citeauthoryear{Gladkova \bgroup \em et al.\egroup
  }{2016}]{gladkova2016analogy}
Anna Gladkova, Aleksandr Drozd, and Satoshi Matsuoka.
\newblock Analogy-based detection of morphological and semantic relations with
  word embeddings: what works and what doesn’t.
\newblock In {\em Proceedings of the NAACL Student Research Workshop}, pages
  8--15, 2016.

\bibitem[\protect\citeauthoryear{Hochreiter and
  Schmidhuber}{1997}]{10.1162/neco.1997.9.8.1735}
Sepp Hochreiter and J\"{u}rgen Schmidhuber.
\newblock Long short-term memory.
\newblock {\em Neural Comput.}, 9(8):1735–1780, nov 1997.

\bibitem[\protect\citeauthoryear{Holyoak \bgroup \em et al.\egroup
  }{1995}]{holyoak1995mental}
Keith~J Holyoak, Paul Thagard, and Stuart Sutherland.
\newblock Mental leaps: analogy in creative thought.
\newblock {\em Nature}, 373(6515):572--572, 1995.

\bibitem[\protect\citeauthoryear{Lenat}{1995}]{lenat1995cyc}
Douglas~B Lenat.
\newblock Cyc: A large-scale investment in knowledge infrastructure.
\newblock {\em Communications of the ACM}, 38(11):33--38, 1995.

\bibitem[\protect\citeauthoryear{Liu \bgroup \em et al.\egroup
  }{2019}]{DBLP:journals/corr/abs-1907-11692}
Yinhan Liu, Myle Ott, Naman Goyal, Jingfei Du, Mandar Joshi, Danqi Chen, Omer
  Levy, Mike Lewis, Luke Zettlemoyer, and Veselin Stoyanov.
\newblock Roberta: {A} robustly optimized {BERT} pretraining approach.
\newblock {\em CoRR}, abs/1907.11692, 2019.

\bibitem[\protect\citeauthoryear{Lu \bgroup \em et al.\egroup
  }{2019}]{lu2019emergence}
Hongjing Lu, Ying~Nian Wu, and Keith~J Holyoak.
\newblock Emergence of analogy from relation learning.
\newblock {\em Proceedings of the National Academy of Sciences},
  116(10):4176--4181, 2019.

\bibitem[\protect\citeauthoryear{Mikolov \bgroup \em et al.\egroup
  }{2013a}]{https://doi.org/10.48550/arxiv.1301.3781}
Tomas Mikolov, Kai Chen, Greg Corrado, and Jeffrey Dean.
\newblock Efficient estimation of word representations in vector space, 2013.

\bibitem[\protect\citeauthoryear{Mikolov \bgroup \em et al.\egroup
  }{2013b}]{mikolov2013linguistic}
Tom{\'a}{\v{s}} Mikolov, Wen-tau Yih, and Geoffrey Zweig.
\newblock Linguistic regularities in continuous space word representations.
\newblock In {\em Proceedings of the 2013 conference of the north american
  chapter of the association for computational linguistics: Human language
  technologies}, pages 746--751, 2013.

\bibitem[\protect\citeauthoryear{Reagan \bgroup \em et al.\egroup
  }{2016}]{DBLP:journals/corr/ReaganMKDD16}
Andrew~J. Reagan, Lewis Mitchell, Dilan Kiley, Christopher~M. Danforth, and
  Peter~Sheridan Dodds.
\newblock The emotional arcs of stories are dominated by six basic shapes.
\newblock {\em CoRR}, abs/1606.07772, 2016.

\bibitem[\protect\citeauthoryear{Ribeiro and
  Forbus}{2021}]{ribeiro2021combining}
Danilo~Neves Ribeiro and Kenneth Forbus.
\newblock Combining analogy with language models for knowledge extraction.
\newblock In {\em 3rd Conference on Automated Knowledge Base Construction},
  2021.

\bibitem[\protect\citeauthoryear{Sanh \bgroup \em et al.\egroup
  }{2019}]{Sanh2019DistilBERTAD}
Victor Sanh, Lysandre Debut, Julien Chaumond, and Thomas Wolf.
\newblock Distilbert, a distilled version of bert: smaller, faster, cheaper and
  lighter.
\newblock {\em ArXiv}, abs/1910.01108, 2019.

\bibitem[\protect\citeauthoryear{Swayamdipta \bgroup \em et al.\egroup
  }{2017}]{https://doi.org/10.48550/arxiv.1706.09528}
Swabha Swayamdipta, Sam Thomson, Chris Dyer, and Noah~A. Smith.
\newblock Frame-semantic parsing with softmax-margin segmental rnns and a
  syntactic scaffold, 2017.

\bibitem[\protect\citeauthoryear{Tomai and Forbus}{2007}]{tomai2007plenty}
Emmett Tomai and Ken Forbus.
\newblock Plenty of blame to go around: a qualitative approach to attribution
  of moral responsibility.
\newblock Technical report, NORTHWESTERN UNIV EVANSTON IL, 2007.

\bibitem[\protect\citeauthoryear{Tversky}{1977}]{tversky1977features}
Amos Tversky.
\newblock Features of similarity.
\newblock {\em Psychological review}, 84(4):327, 1977.

\bibitem[\protect\citeauthoryear{Ushio \bgroup \em et al.\egroup
  }{2021a}]{DBLP:journals/corr/abs-2105-04949}
Asahi Ushio, Luis~Espinosa Anke, Steven Schockaert, and Jos{\'{e}}
  Camacho{-}Collados.
\newblock {BERT} is to {NLP} what alexnet is to {CV:} can pre-trained language
  models identify analogies?
\newblock {\em CoRR}, abs/2105.04949, 2021.

\bibitem[\protect\citeauthoryear{Ushio \bgroup \em et al.\egroup
  }{2021b}]{ushio-etal-2021-bert}
Asahi Ushio, Luis Espinosa~Anke, Steven Schockaert, and Jose Camacho-Collados.
\newblock {BERT} is to {NLP} what {A}lex{N}et is to {CV}: Can pre-trained
  language models identify analogies?
\newblock In {\em Proceedings of the 59th Annual Meeting of the Association for
  Computational Linguistics and the 11th International Joint Conference on
  Natural Language Processing (Volume 1: Long Papers)}, pages 3609--3624,
  Online, August 2021. Association for Computational Linguistics.

\end{thebibliography}
